\documentclass[letterpaper, 10 pt, conference]{ieeeconf}  % Comment this line out if you need a4paper

\IEEEoverridecommandlockouts                              % This command is only needed if 
\usepackage{hyperref}
\usepackage{graphicx}
\usepackage{amsmath}
\usepackage{cite}
\usepackage{algorithm}
\usepackage{algpseudocode}
\usepackage{multirow}

\title{\LARGE \bf{RoboChop: Autonomous Framework for Fruit and Vegetable Chopping Leveraging Foundational Models}}

\author{Atharva Dikshit$^{*1}$, Alison Bartsch$^{*1}$, Abraham George$^{1}$, and Amir Barati Farimani$^{1}$
\thanks{$^{*}$These authors made an equal contribution to this work}
\thanks{$^{1}$With the Department of Mechanical Engineering,
        Carnegie Mellon University \tt\small \{adikshit, abartsch, aigeorge, afariman\} @andrew.cmu.edu}}

\begin{document}

\maketitle
\thispagestyle{empty}
\pagestyle{empty}

% Previous Header info: 

% \documentclass{article}

% % Language setting
% % Replace `english' with e.g. `spanish' to change the document language
% \usepackage[english]{babel}

% % Set page size and margins
% % Replace `letterpaper' with`a4paper' for UK/EU standard size
% \usepackage[letterpaper,top=2cm,bottom=2cm,left=3cm,right=3cm,marginparwidth=1.75cm]{geometry}

% % Useful packages
% \usepackage{amsmath}
% \usepackage{graphicx}
% \usepackage[colorlinks=true, allcolors=blue]{hyperref}

\begin{abstract}
With the goal of developing fully autonomous cooking robots, developing robust systems that can chop a wide variety of objects is important. Existing approaches focus primarily on the low-level dynamics of the cutting action, which overlooks some of the practical real-world challenges of implementing autonomous cutting systems. In this work we propose an autonomous framework to sequence together action primitives for the purpose of chopping fruits and vegetables on a cluttered cutting board. We present a novel technique to leverage vision foundational models SAM and YOLO to accurately detect, segment, and track fruits and vegetables as they visually change through the sequences of chops, finetuning YOLO on a novel dataset of whole and chopped fruits and vegetables. In our experiments, we demonstrate that our simple pipeline is able to reliably chop a variety of fruits and vegetables ranging in size, appearance, and texture, meeting a variety of chopping specifications, including fruit type, number of slices, and types of slices. 

\end{abstract}

\section{Introduction}

As humans, the task of slicing fruits and vegetables on a cutting board is a common and relatively mundane task. When cooking a recipe, we may have several goals for the types of slices we would like to cut of the various objects on the cutting board (i.e. evenly chopped, long and thin strips, diced, etc.). We sequence together various skills to achieve these different target states for the fruits and vegetables we are cutting, easily pushing some objects out of the way to clear a small workspace for each slicing action. However, for autonomous robotic systems, this scenario is challenging for a few reasons. Firstly, the mechanics of cutting deformable fruits and vegetables are complicated, making the cutting action itself difficult to execute and generalize to a variety of objects \cite{mu2019slicing, heiden2021disect, long2014cutting, jamdagni2021}. Second, fruits and vegetables will appear different between their whole and chopped forms, making identifying and tracking sliced objects throughout the duration of the task challenging. Third, this task contains a difficult reasoning problem: planning how to sequence actions together to execute a variety of goals.

As the robotics field works towards developing autonomous cooking robots \cite{liu2022, xiao2022robotic, sochacki2021} and household assistive robots \cite{li2021igibson, wake2021, kazhoyan2021}, developing systems that can handle the complex variations within this task of chopping fruits and vegetables on a cutting board is necessary. In this work, we present a vision-based framework that reliably chops various fruits and vegetables on a cluttered workspace, fully autonomously leveraging  vision foundational models YOLO \cite{yolov8_ultralytics} and SAM \cite{kirillov2023segment} and a library of action primitives. Existing work within the cooking and object slicing domain tends to either be at the highest-level, planning high level cooking actions \cite{takata2022graph, sakib2022cooking, delpreto2022actionsense}, or the lowest level, planning the slicing motion itself \cite{zhang2019, mu2019slicing}. In this work, we focus on the mid-level task of stringing together action primitives to autonomously slice objects. To this end, we do not focus on the intricacies and dynamics of the slice action itself, and instead focus on the additional skills necessary to result in a robust autonomous chopping pipeline. Our key contributions are as follows:

\begin{itemize}
    \item We present a framework to successfully accomplish the cluttered cutting board task by autonomously sequencing together a series of low-level action primitives.

    \item We discuss in detail a practical approach to leveraging Segment Anything Model (SAM) for real-world robotic tasks by combining SAM with YOLO to iteratively combine relevant partial masks of the objects of interest.

    \item We present a new object detection dataset for 3 classes of fruits and vegetables to detect both whole and chopped-up versions of the fruits and vegetables. To our knowledge, this is the first publicly available dataset of its kind, and we plan to continue to expand the food classes it contains.
\end{itemize}

\begin{figure}
      \centering
     \includegraphics[width=1.0\linewidth]{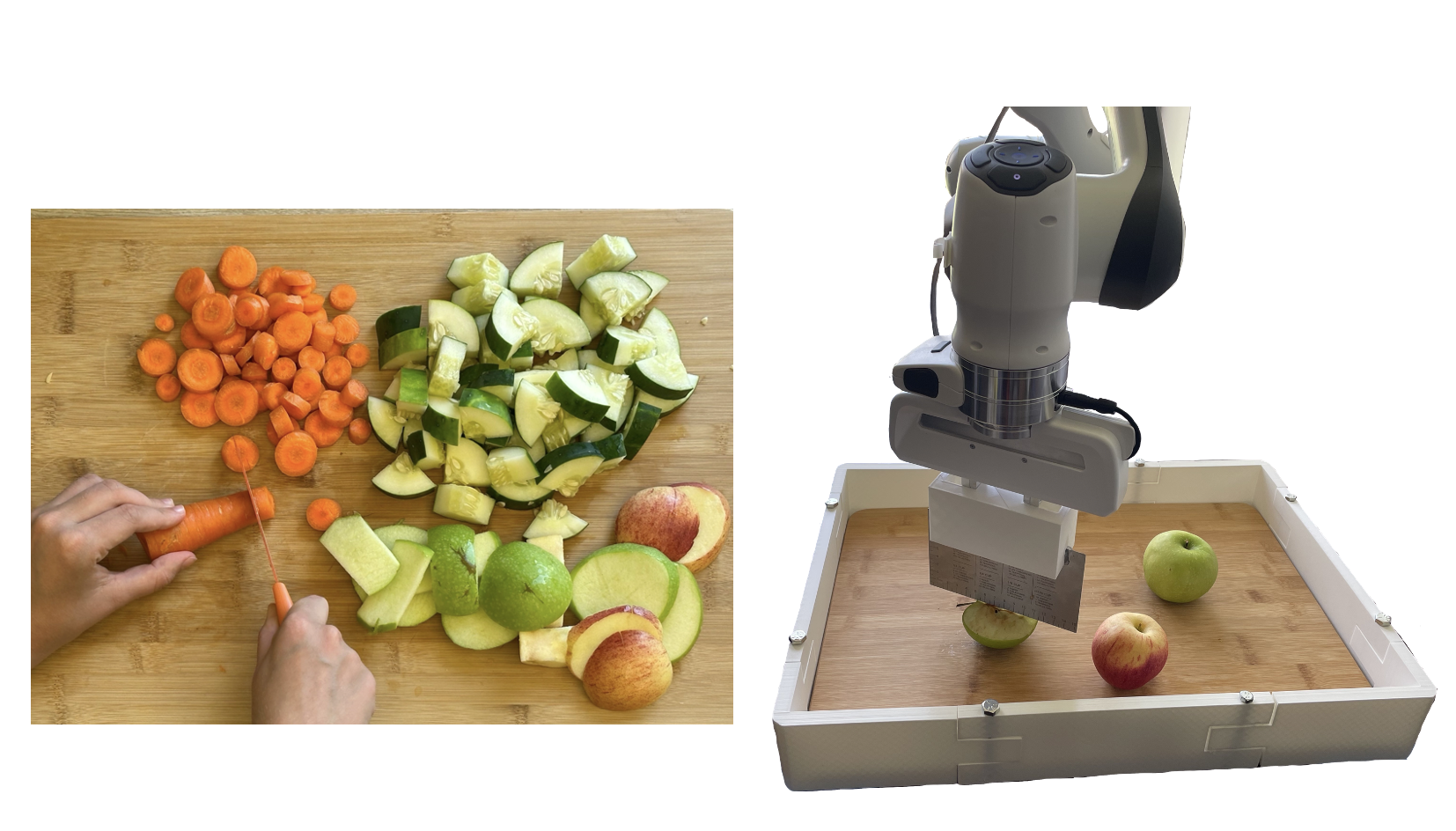}
      \caption{\label{fig:pipeline}\textit{Left: } Task motivation of a cluttered cutting board. \textit{Right: } Proposed autonomous slicing system within a cluttered cutting board scenario.}
      \label{figurelabel}
  \end{figure}

\section{Related works}

\textbf{Visual Foundational Models: } There has been much work in the computer vision community to develop large-scale general models trained on a large quantity of data known as foundation models that can be leveraged for a variety of applications. YOLO \cite{redmon2016you} presented a network that accomplished real-time object detection and has since been improved upon over the years \cite{redmon2017yolo9000, redmon2018yolov3, bochkovskiy2020yolov4}. YOLO has become the primary object detection model for robotic pipelines, with uses ranging from quadruped robots \cite{norizan2023object} to cable insertion tasks \cite{mou2022pose}. More recently, SAM \cite{kirillov2023segment} demonstrates a remarkable ability to provide detailed segmentation of cluttered scenes. However, SAM does not provide any labels for the objects it segments and often requires bounding box or point-based queries to provide the most detailed segmentation of objects, making it difficult to use in practice in fully autonomous systems without using a secondary vision model. Alternate models have garnered much success within the robotics field, particularly CLIP \cite{radford2021learning}, which jointly trains an image encoder and text encoder to predict image and text pairings. CLIP has enabled the incorporation of language-conditioned goals in a variety of robotics work. For example, CLIPORT  \cite{shridhar22a} is a language-conditioned imitation learning framework that leveraged CLIP to incorporate language into policy training. R3M \cite{nair2022r3m} is a manipulation-specific visual foundational model that was trained on the Ego4D dataset \cite{grauman2022ego4d} consisting of egocentric videos of humans performing a wide variety of tasks. R3M, or fine-tuned R3M has been used as the perception module in a variety of manipulation works \cite{wang2023manipulate, chane2023learning}.

\textbf{Robotic Slicing: } Existing works addressing robotic slicing tend to focus on the mechanics of the cutting action itself. \cite{heiden2021disect} explores the complex mechanics of cutting soft materials and presents a fully differentiable simulation to model cutting. \cite{ikegami2020} leverages a simulation environment to train a neural network to predict the state change of an object after being cut. \cite{jamdagni2021} models the cutting mechanics with even more detail, using the finite element method (FEM) to model the forces generating the fracture and deformation of the objects being sliced. The more practical implementations of robotic slicing present various approaches, typically leveraging a combination of force and vision information. \cite{mu2019slicing} presents a control framework for a 2DOF robotic arm, specifically by decomposing the cutting action into the sub-actions of press, push, and slice. Similarly, \cite{long2014cutting} presents a force and vision control strategy to model and control the complex cutting dynamics. \cite{zhang2019} leverages multimodal sensor data for robust cutting, decomposing the slice action into a series of expected events, such as hitting the object, scraping the object, and slicing. \cite{sawhney2021playing} explicitly explores how to model the material properties of the food being cut, and argues the importance of multimodal sensor information.

\textbf{Kitchen Task Planning and Understanding: } Other works within the autonomous cooking domain work at the highest level, planning sequences of actions to complete long-horizon cooking tasks instead of the practical execution of these actions in the real world. \cite{blodow2011} autonomously generates 3D semantic maps while a robot performs kitchen tasks and explores the environment. \cite{zhu2017} developed a verbalization system that provides a natural language narration as a robot explores a simulated kitchen environment. \cite{sakib2022cooking} presents a knowledge network for cooking that outputs a task tree for completing recipes. \cite{takata2022graph} presents a graph-based approach for planning cooking actions from a recipe. \cite{delpreto2022actionsense} creates a multimodal dataset of humans performing kitchen tasks to be leveraged for imitation learning-based cooking systems. \cite{wang2020too} presents a planning framework for multi-agent collaboration for kitchen tasks, once again with experiments in simulation. While each of these techniques explores interesting aspects of cooking tasks, they consistently lack real-world experimental validation, which is crucial for building systems that address some of the ongoing real-world and practical challenges within the cooking domain.

\section{Methodology}

\begin{figure*}
      \centering
     \includegraphics[width=0.93\linewidth]{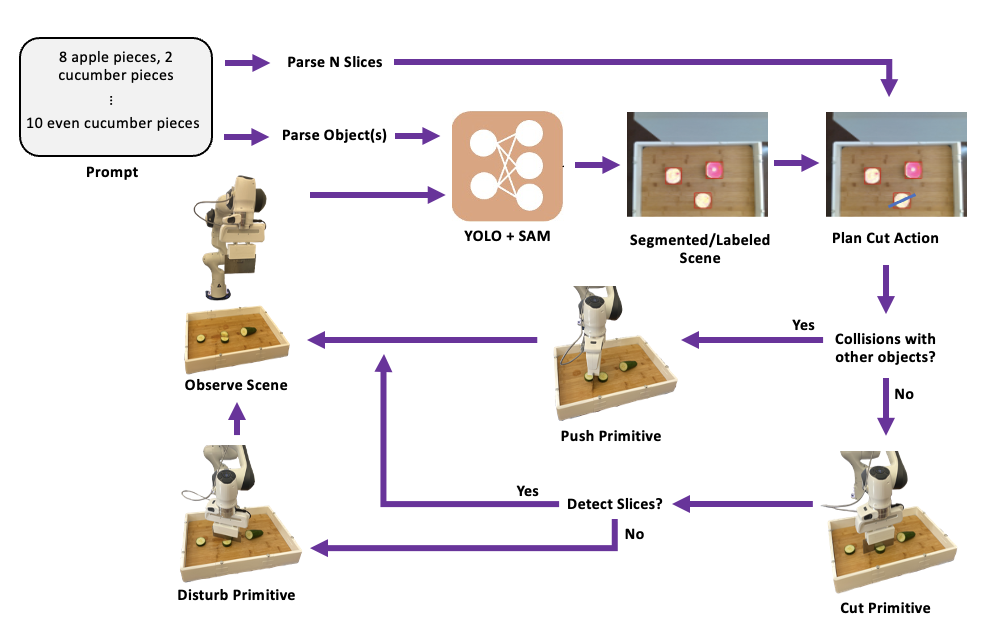}
      \caption{\label{fig:pipeline}An overview of the system pipeline. The text-based prompts are parsed to identify the goal conditions, particularly the number of slices and cutting style for each fruit/vegetable to be cut. This information is fed to the vision system to label and segment the scene, extracting the centroid positions of each object on the cutting board. With the goal conditions, the planner selects the objects to cut in order of largest segmentation mask area to smallest. The cutting action is planned, generating the object to cut and rotation of the blade. This planned cut is checked for collisions with objects other than the target to slice. If there are expected collisions, the push primitive is executed to separate the objects, and the scene is re-observed and segmented. When no collisions are detected, the cut primitive is executed, and the scene is re-observed. If the number of objects observed is less than that expected (with the assumption that the cut action was successful), the disturb primitive is executed, because the assumption is that the two slices are too close together for the vision system to correctly segment them. }
      \label{figurelabel}
  \end{figure*}

In this section, we provide a detailed explanation of all the components of the autonomous chopping system. A visual overview of the pipeline is shown in Figure \ref{fig:pipeline}. The vision system combining YOLO with SAM is detailed in section \ref{sec:vision}, an explanation of the action primitives is provided in section \ref{sec:primitives}, and a discussion of the action sequencing logic is in section \ref{sec:loop}.

\subsection{Vision System}
\label{sec:vision}

\begin{figure*}
      \centering
     \includegraphics[width=0.8\linewidth]{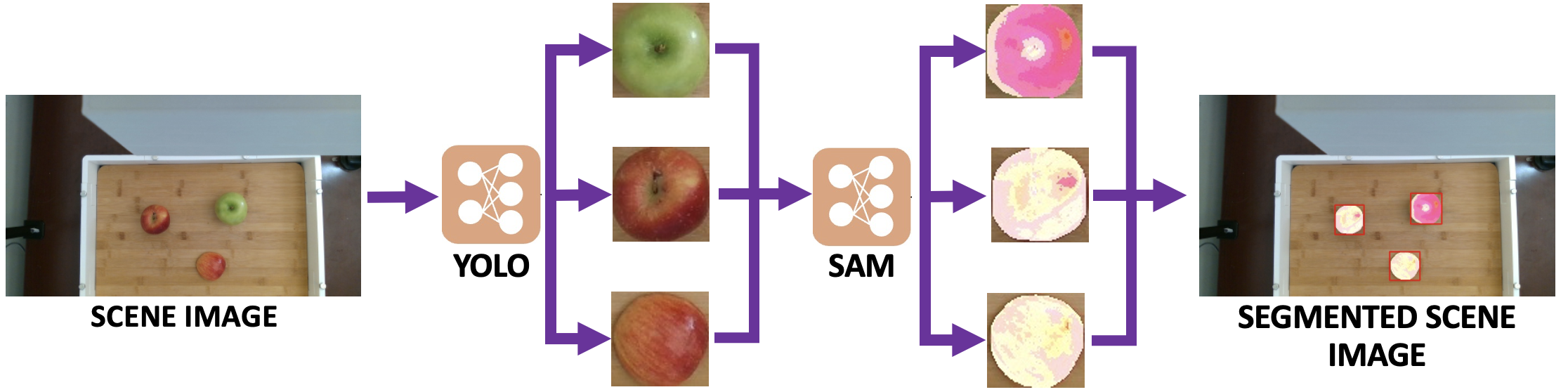}
      \caption{\label{fig:vision_pipeline}An overview of the entire vision pipeline. The wrist-mounted camera captures the scene at the robot's observation pose. This image is then passed through YOLOv8 for fast object detection. Each object is then cropped with its bounding box and passed through SAM to provide an accurate segmentation of the objects in the scene. Refer to Figure \ref{fig:sam_yolo} for details on how the SAM masks are combined for each cropped region. These masks are then combined with the original image to provide a full scene segmentation of the fruits and vegetables.}
      \label{figurelabel}
  \end{figure*}

\begin{figure*}
      \centering
     \includegraphics[width=0.85\linewidth]{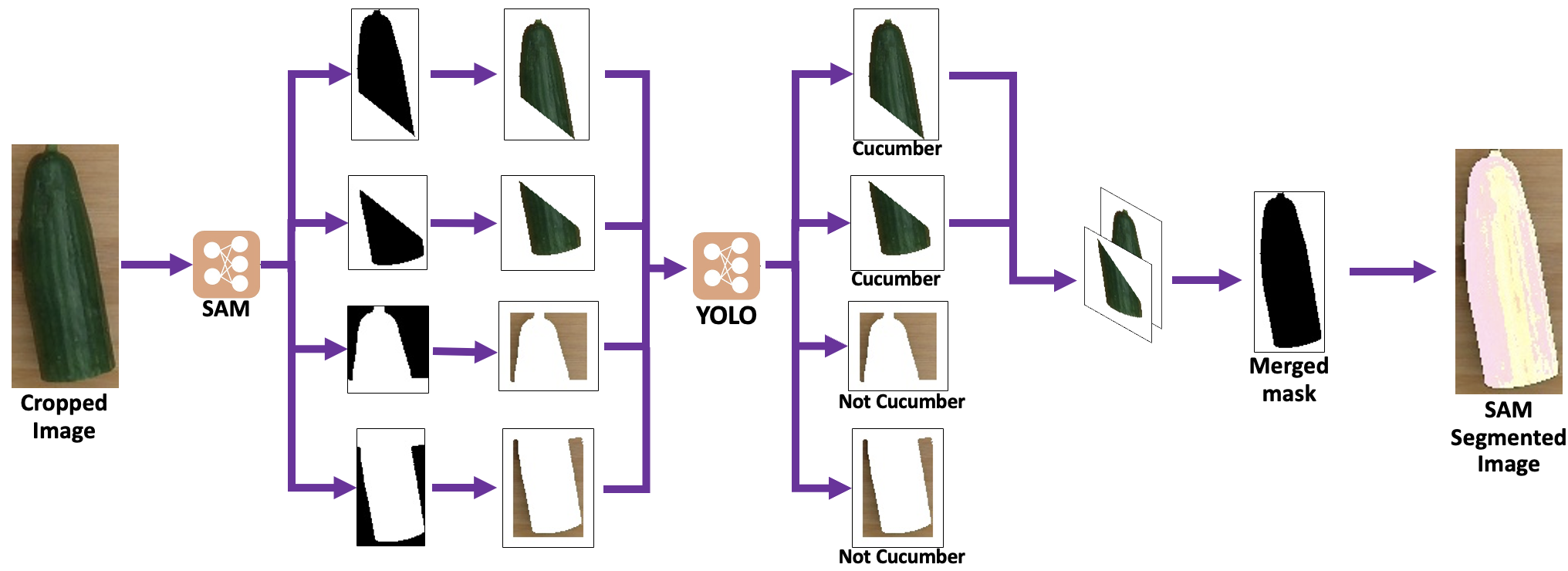}
      \caption{\label{fig:sam_yolo}An overview of the SAM segmentation of a cropped image. The cropped image generated using the YOLO bounding box coordinates is passed through  SAM to get a list of segmentation masks. The binary mask are then merged with the cropped image to display only the part of the cropped image that has been segmented by SAM. These images are then processed by YOLO to output the predicted labels. Finally, if the labels match that of the target object, the masks are merged together to output a single binary mask of the target object. }
      \label{figurelabel}
  \end{figure*}

% Abraham Version
Our vision system attempts to segment the foods we wish to cut using the powerful pre-trained segmentation model SAM (Segment Anything Model) by Meta Research team \cite{kirillov2023segment}. However, the segmentation masks output by SAM are unlabeled, which limits the model's utility for our task (segmenting out specific fruits and vegetables, both sliced and unsliced) and for robotic tasks in general, where both segmentation and identification are often required. To overcome this limitation we used another machine learning model to assist SAM, allowing us to identify and segment objects of interest in the scene. Specifically, we employed a YOLO v8 model from Ultralytics \cite{yolov8_ultralytics}. Using this pre-trained model, we were able to construct bounding boxes around uncut fruits, but the system failed to identify sliced fruits. To address this problem, we fine-tuned the YOLO model on a dataset of sliced and uncut fruits and vegetables. As the problem we are tackling is quite novel, there were few datasets that contained both cut and uncut fruits and vegetables with bounding box labels. To that end, we created our own custom dataset of both cut and uncut apples, cucumbers and carrots. We plan to increase the number of classes within our dataset for future cooking tasks. Apart from collecting our own dataset, we further added a few augmentation techniques to the images to make our fine-tuned YOLO model more robust. The augmentation techniques were generated by Roboflow and randomly applied to the images based on a range that we specified. The augmentation techniques we used are as follows:
\begin{enumerate}
    \item Flip (Horizontal and Vertical)
    \item Saturation (between -25\% and +25\%)
    \item  Blur (Up to 2.5px)
    \item Noise (Up to 5\% of pixels)
\end{enumerate}
Once trained on this new dataset, our YOLO model could reliably segment fruit, both cut and whole. Although the YOLO model is sufficient for labeling instances of fruits, it only provides a bounding box of the fruit instance, not a mask. A mask is vital for proper planning of the cutting action, especially if the food has a high degree of curvature or is orientated such that it only occupies a small portion of the bounding box (ie. a banana, a cantaloupe slice, or a carrot lying diagonally). To get this mask, we apply SAM to the section of the image inside of the YOLO bounding box, resulting in a collection of masks, one (or more) being the mask of the food. To identify which masks belong to the object and which are the background, we apply YOLO to each segmentation (placed on a white background). The segments that are identified as the food in question are then combined to create a final food mask. This is necessary, as there is no guarantee what kinds of masks SAM will generate, and it can often segment a single object into multiple smaller masks that need to be combined. Although identifying segments of food rather than whole foods increases the difficulty of the identification task, since our YOLO model is also trained on cut fruits, this challenge does not pose a significant limitation. A visualization of the vision pipeline is shown in Figure \ref{fig:vision_pipeline}, with a visualization of the iterative mask selection shown in Figure \ref{fig:sam_yolo}. Once the food is segmented, the center of the food is identified as the average x and y coordinate of the segmentation mask, which is then transformed into the robot world coordinate frame. This centroid, along with the mask, is then sent to the cutting primitive, which uses them to determine the exact location and orientation of the cut.

\subsection{Action Primitive Library}
\label{sec:primitives}

We provide the system with a small library of skills to successfully achieve the chopping task. These skills include \textit{cut}, \textit{disturb}, \textit{observe}, \textit{detect collision} and \textit{push}. For this work, we chose to focus on the mid-level complexities of chopping various fruits and vegetables, and thus made a series of assumptions about the cutting action itself. We define the \textit{cut} skill as applying a large force down in the z-axis at a specified position and blade rotation, without any sliding or rocking back and forth. We found in practice this simplifying assumption worked well when applying sufficient downward force. The cut position is defined as the predicted centroid of the target fruit, and the blade rotation is determined based on the user input's slice specification. If the user would like evenly sized pieces, the blade is rotated such that it is perpendicular to the longest diameter in the object's segmentation mask. If the user would like longer pieces, the blade is rotated parallel to this longest line. This simple heuristic allows for variation in chopping style that is typically required in common recipes. An additional assumption within our pipeline is that the cut action is always successful. Thus, we have developed the \textit{disturb} skill, such that if the vision system does not detect the expected number of pieces in the scene after a \textit{cut} action is executed, the robot returns to the previous cut pose and applies a random end-effector rotation to perturb the scene, separating the sliced pieces that may still be stuck together. For the \textit{observe} skill, the end-effector moves to a specified overhead pose to view the entire workspace and use the vision system to segment and label all objects in the scene. The \textit{detect collision} skill is called before each \textit{cut} action is executed to determine if any objects in the scene that are not the target of the planned cut will come into contact with the blade. If so, the \textit{push} skill is called to push the interfering object out of the way. The specific vector along which the robot pushes the interfering object is calculated based on the center of mass predictions of the target object and the interfering object. The skills within the skill library could be modified or improved upon with alternative primitive collection techniques, such as imitation learning, or training a policy from scratch with reinforcement learning. However, for the purpose of this work, these pre-defined skills were sufficient.

\subsection{Autonomous Loop}
\label{sec:loop}

We propose a heuristic-based autonomous loop to sequence between the action primitives in the skill library. The system logic relies on two key assumptions: (1) the cut action is successful and produces two pieces of the cut object, and (2) all objects will remain in the scene for the duration of execution. These assumptions are necessary, because there are circumstances where an object is cut, but the two pieces are not accurately detected because they remain somewhat stuck together. By assuming that all cut actions are successful, and all objects will remain in the scene, if the detected number of objects after a cut action are not what we expected, we can conclude that the slices are stuck together, and the disturb action primitive will fix this. Figure \ref{fig:pipeline} visualizes the system pipeline, while Algorithm \ref{alg:logic} provides the pseudocode for action primitive selection.

\begin{algorithm}
\caption{\label{alg:logic}System Pipeline Pseudocode}
\label{figurelabel}
\begin{algorithmic}
% \Require $skills = {observe(), cut(), push(), disturb()}$
\\
\textbf{Input:} $N_{target}$ 
\\
\textbf{Input:} $Class_{target}$
\\
\textbf{Input:} $skills = observe(), plan\_cut(), cut(), disturb(), $

$push(), check\_collisions()$
\\
$N_{obs}, Obj_{dict} = skills.observe(Class_{target})$
\\
$count = N_{obs}$
\While {$N_{target} \geq N_{obs}$} 

    \If{$count \neq N_{obs}$} \Comment{If new slice not detected}
        \State $skills.disturb()$
        
        \State $N_{obs}, Obj_{dict} = skills.observe(Class_{target})$
    \Else{} \Comment{Plan and execute cut action}
        \State $com, angle = skills.plan\_cut(Obj_{dict})$
        \State $col = skills.check\_collisions(com, angle, Obj_{dict})$

        \For {$idx$ \text{in} $col$}
            \State $skills.push(Obj_{dict} [idx])$
        \EndFor
        \State $N_{obs}, Obj_{dict} = skills.observe(Class_{target})$
        \State $count = skills.cut(count, com, angle)$
        \State $N_{obs}, Obj_{dict} = skills.observe(Class_{target})$
    \EndIf

\EndWhile
\end{algorithmic}
% \label{sec:logic}
\end{algorithm}

\section{Experimental Setup Details}

\begin{figure}
      \centering
     \includegraphics[width=0.7\linewidth]{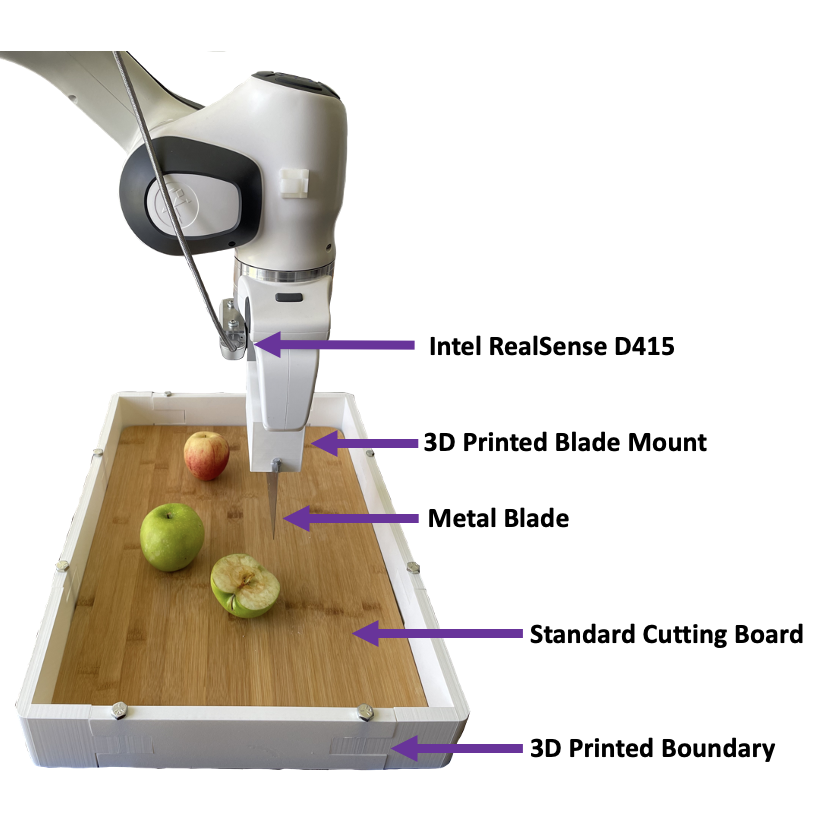}
      \caption{\label{fig:hardware}A visualization of our hardware workspace. We 3D printed a tool mount for the Franka robot to grasp the cutting tool. We additionally 3D printed a boundary wall surrounding the cutting workspace to prevent objects from rolling off following a rapid cutting action.}
      \label{figurelabel}
  \end{figure}

We test and evaluate our proposed pipeline entirely in real-world experiments with a 7DOF Franka robot arm \cite{franka2021}, with our control module built off of frankapy \cite{zhang2020modular}. The robot is equipped with a wrist-mounted Intel RealSense D415 camera, and a 3D printed blade tool. This blade tool is designed to be easily interchangeable by the robot, for future tasks that require multiple tools. The blade is designed for a chopping action applying force downward, as opposed to a standard knife cutting action which may involve lateral motion as well. Additionally, we provide a 3D printed barrier around the cutting board to prevent pieces from rolling off the workspace. A detailed visualization of our experimental setup is shown in Figure \ref{fig:hardware}.

\section{Results}

We conduct a series of experiments evaluating the robustness of the pipeline to variations in fruit and vegetable types, appearances, poses and scene clutter. First, in \ref{exp1} we evaluate each component of our system at the task of a multi-class cluttered environment. Next, in \ref{exp2} we evaluate the reliability of the chop action primitive itself, varying the cut type heuristic and fruit/vegetable class. Finally, in \ref{exp3} we evaluate our system on a set of multi-step, multi-class cluttered environment tasks to determine how well our pipeline can recover from partial failure, and how reliably it can meet target goals.

\subsection{Evaluation of Vision, Planning and Action Execution}
\label{exp1}

\begin{table}[]
\centering
\begin{tabular}{@{\extracolsep{\fill}}lllll}
        \hline
         \textbf{ } & \textbf{ } & \multicolumn{2}{c}{\textbf{Success Rate}}\\
         \textbf{ } & \textbf{ } & \textbf{\textit{Two-Class}} & \textbf{\textit{Three-Class}}\\ 
        \hline
        \hline
        \multirow{2}{*}{\textbf{Vision}} & \textbf{\textit{YOLO}} & 76\% (OL 93.4\%) & 80\% (OL 92.9\%) \\
         & \textbf{\textit{SAM}} & 92\% (OL 97.5\%) & 90\% (OL 90\%) \\
        \hline
        \multirow{2}{*}{\textbf{Planner}} & \textbf{\textit{Cut}} & 100\% & 90\%\\
         & \textbf{\textit{Collision}} & 100\% & 80\% \\
        \hline 
        \multirow{3}{*}{\textbf{Action Execution}} & \textbf{\textit{Cut}} & 76.5\% & 66.7\% \\
         & \textbf{\textit{Collision}} & 69.2\% & 50\% \\
         & \textbf{\textit{Disturb}} & 66.7\% & - \\
        \hline 
    \end{tabular}
\caption{\label{tab:firstexp} \textbf{The success rate of each component of the system with two and three fruit/vegetable classes and slices varying in size.} There were 25 randomized trials for the two-class experiments, and 10 randomized trials for the three-class experiments. The object-level (OL) success in brackets reports the success rate for each object. In the three-class experiments, the disturb action primitive was never called.} \label{table1}
\end{table}

To evaluate each component of the pipeline on a cluttered scene, we designed a procedural scene parameter generator to ensure scene setups were entirely randomized, and not biased by our object selection. The number of objects initialized on the cutting board ranged between 1 and 10, with each object being randomly assigned a fruit/vegetable class, and a slice size ranging from whole to $\frac{1}{8}$. The target object to cut was also randomly determined. We conducted a first set of 25 experiments with two possible object classes - apple and cucumber, and an additional 10 experiments with three possible object classes - apple, cucumber and carrot, all shown in Table \ref{tab:firstexp}. The purpose of these experiments is to evaluate each system component on a cluttered scene, where the goal is to chop the target object in half. For each system we define a run as successful if the component succeeded with all objects in the scene. We report an additional object-level success rate for the vision systems as there are frequent scenarios where nearly all objects in the scene are successfully detected and segmented, and the system is able to successfully perform the cutting task. Averaging between the two class (Table \ref{exp1}) and three class experiments (Table \ref{exp2}, the YOLO system achieves a 77\% overall success rate, the SAM system achieves a 91\% overall success rate, cut planning is 97\% successful, collision detection is 94\% successful, and the chop action itself is 74\% successful.

\subsection{Chopping Single Object}
\label{exp2}

The goal of this experiment is to evaluate the quality of the simplified chopping action, specifically to investigate how it may vary in success depending on the fruit/vegetable and chop heuristic. We conducted 5 experimental runs for each object and chop heuristic combination, with 30 experiments total. For each experiment, a single instance of that object was placed in the center of the cutting board, varying in slice size from a whole object to a $\frac{1}{8}$ slice. The chop heuristic even corresponds to rotating the blade perpendicular to the longest line in the mask, to encourage even slices. The chop heuristic long corresponds to rotating the blade parallel to the longest line in the mask, to encourage long and thin slices. The results in Table \ref{table3} show that the even heuristic cut action is more reliable, particularly on long and thin objects, such as cucumbers and carrots. This is an intuitive result, as in our simplified cutting action the robot applies a downward force. However, when cutting a cylindrical object, this downward force can easily roll the object instead of chopping it. This suggests a need for a more sophisticated chop primitive when cutting long and very thin objects, such as carrots, into strips.

\begin{table}[]
\centering
\begin{tabular}{@{\extracolsep{\fill}}lll}
        \hline
         \textbf{ } & \textbf{ } & \textbf{Success Rate} \\ 
        \hline
        \hline
        \multirow{2}{*}{\textbf{Apple}} & Even & 100\% \\
         & Long & 100\% \\
        \hline
        \multirow{2}{*}{\textbf{Cucumber}} & Even & 100\% \\
         & Long & 80\% \\
        \hline 
        \multirow{3}{*}{\textbf{Carrot}} & Even & 80\% \\
         & Long & 40\% \\
        \hline 
    \end{tabular}
\caption{\label{tab:failure} \textbf{The success rate for a single chop varying the object class and cut type.}} \label{table3}
\end{table}

\begin{table}[ht]
    \centering
    \begin{tabular}{@{\extracolsep{\fill}}llllll}
    % \begin{tabular}{@{\extracolsep{\fill}}|c|c|c|c|c|c}
    % \begin{tabular}{@{\extracolsep{\fill}{|C|C|C|C|C|C|}
        \hline
         \textbf{Experiment} & \multicolumn{2}{c}{\textbf{Initial \#}} & \multicolumn{2}{c}{\textbf{Target \#}} & \textbf{Success?} \\ 
         \textbf{ } & \textbf{A} & \textbf{C} & \textbf{A} & \textbf{C} & \textbf{} \\
        \hline
        \hline
        1 & 4 & 2 & 8 & 3 & No \\
        2 & 2 & 2 & 3 & 4 & Yes \\
        3 & 1 & 2 & 3 & 4 & Yes \\
        4 & 3 & 0 & 8 & 0 & Yes \\
        5 & 0 & 4 & 0 & 8 & Yes \\
        6 & 0 & 5 & 0 & 8 & No \\
        7 & 2 & 1 & 4 & 3 & Yes \\
        8 & 0 & 5 & 0 & 7 & Yes \\
        9 & 2 & 0 & 6 & 0 & No \\
        10 & 2 & 0 & 7 & 0 & Yes \\
        \hline 
    \end{tabular}
    \caption{\label{tab:failure} \textbf{The results of 10 mutli-object cluttered experiments with varying target slices.} 'A' represents apples and 'C' represents cucumbers. The initial number represents the number of pieces of that class initialized in the scene. The target number represents the target number of pieces to be in the scene after the sequence of cutting actions.}
    \label{table4}
\end{table}

\subsection{Chopping Multi-Object Scenes}
\label{exp3}

\begin{figure}
      \centering
     \includegraphics[width=0.85\linewidth]{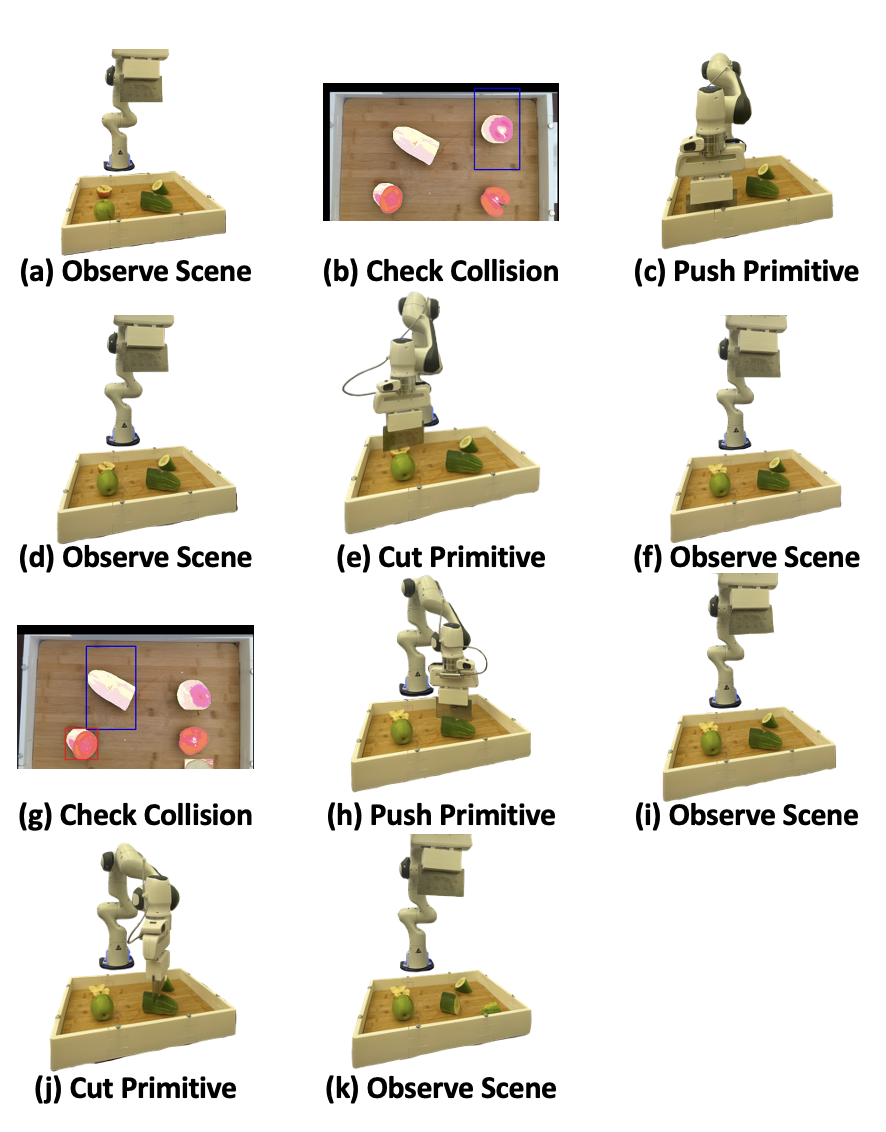}
      \caption{\label{fig:sequence}A visualization of the action sequence for experiment 2 of the cluttered scene with apples and cucumbers.}
      \label{exp}
  \end{figure}

We deployed our system on 10 randomized scenes with varying the number of objects, classes, sizes, and target number of cuts. The goal of these experiments was to observe how well the system can meet target chop specifications, as well as how resilient the pipeline is to small failures. The results of the experiments are shown in Table \ref{table4}, and the pipeline of one experiment is visualized in Figure \ref{exp}. Overall, the system was fully successful at 70\% of the tasks. Experiments 1, 6, and 9 failed on subsequent chopping actions when the object rolled out from underneath the blade. Experiment 10 was a success, and is an experiment that exemplifies the necessity of including the disturb primitive. After the first chop action, the apple failed to fully separate in half, making it difficult for the vision system to detect the expected number of slices. The disturb primitive was called to return to the chop location and was able to successfully separate the pieces, detect the pieces, and continue with the chopping action sequence to reach the target number of slices. These experiments demonstrate our system is robust to partial failures, particularly in the vision system.

\section{Conclusion}
In this work, we presented a simple yet effective robotic framework for chopping varied fruits and vegetables autonomously. Beyond the cooking domain, we believe our vision system leveraging YOLO and SAM could be beneficial for a wide range of robotic tasks that could be improved with quality object masking. However, despite high success rates for the vision system and the chopping action, this pipeline exhibits a few limitations. Particularly the assumption that the cutting action is always successful, and the usage of workspace barriers to restrict the system. Additionally, the inference time of SAM may limit scaling this vision system for very large number of objects. Future work should focus on enhancing the cut action success detector, incorporate tool changing to pick and place objects, and optimizing SAM's inference time to handle larger numbers of objects. Our study contributes towards the development of autonomous cooking robots and identifies challenges that need to be addressed.

\newpage
\bibliographystyle{IEEEtran}
\bibliography{ref}

% Generated by IEEEtran.bst, version: 1.14 (2015/08/26)
\begin{thebibliography}{10}
\providecommand{\url}[1]{#1}
\csname url@samestyle\endcsname
\providecommand{\newblock}{\relax}
\providecommand{\bibinfo}[2]{#2}
\providecommand{\BIBentrySTDinterwordspacing}{\spaceskip=0pt\relax}
\providecommand{\BIBentryALTinterwordstretchfactor}{4}
\providecommand{\BIBentryALTinterwordspacing}{\spaceskip=\fontdimen2\font plus
\BIBentryALTinterwordstretchfactor\fontdimen3\font minus
  \fontdimen4\font\relax}
\providecommand{\BIBforeignlanguage}[2]{{%
\expandafter\ifx\csname l@#1\endcsname\relax
\typeout{** WARNING: IEEEtran.bst: No hyphenation pattern has been}%
\typeout{** loaded for the language `#1'. Using the pattern for}%
\typeout{** the default language instead.}%
\else
\language=\csname l@#1\endcsname
\fi
#2}}
\providecommand{\BIBdecl}{\relax}
\BIBdecl

\bibitem{mu2019slicing}
X.~Mu, Y.~Xue, and Y.-B. Jia, ``Robotic cutting: Mechanics and control of knife
  motion,'' in \emph{2019 International Conference on Robotics and Automation
  (ICRA)}, 2019, pp. 3066--3072.

\bibitem{heiden2021disect}
E.~Heiden, M.~Macklin, Y.~Narang, D.~Fox, A.~Garg, and F.~Ramos, ``Disect: A
  differentiable simulation engine for autonomous robotic cutting,'' 2021.

\bibitem{long2014cutting}
P.~Long, W.~Khalil, and P.~Martinet, ``Force/vision control for robotic cutting
  of soft materials,'' in \emph{2014 IEEE/RSJ International Conference on
  Intelligent Robots and Systems}, 2014, pp. 4716--4721.

\bibitem{jamdagni2021}
P.~Jamdagni and Y.-B. Jia, ``Robotic slicing of fruits and vegetables: Modeling
  the effects of fracture toughness and knife geometry,'' in \emph{2021 IEEE
  International Conference on Robotics and Automation (ICRA)}, 2021, pp.
  6607--6613.

\bibitem{liu2022}
J.~Liu, Y.~Chen, Z.~Dong, S.~Wang, S.~Calinon, M.~Li, and F.~Chen, ``Robot
  cooking with stir-fry: Bimanual non-prehensile manipulation of semi-fluid
  objects,'' \emph{IEEE Robotics and Automation Letters}, vol.~7, no.~2, pp.
  5159--5166, 2022.

\bibitem{xiao2022robotic}
C.~Xiao and L.~Zhao, ``Robotic chef versus human chef: The effects of
  anthropomorphism, novel cues, and cooking difficulty level on food quality
  prediction,'' \emph{International Journal of Social Robotics}, vol.~14,
  no.~7, pp. 1697--1710, 2022.

\bibitem{sochacki2021}
G.~Sochacki, J.~Hughes, S.~Hauser, and F.~Iida, ``Closed-loop robotic cooking
  of scrambled eggs with a salinity-based ‘taste’ sensor,'' in \emph{2021
  IEEE/RSJ International Conference on Intelligent Robots and Systems (IROS)},
  2021, pp. 594--600.

\bibitem{li2021igibson}
C.~Li, F.~Xia, R.~Mart{\'\i}n-Mart{\'\i}n, M.~Lingelbach, S.~Srivastava,
  B.~Shen, K.~Vainio, C.~Gokmen, G.~Dharan, T.~Jain \emph{et~al.}, ``igibson
  2.0: Object-centric simulation for robot learning of everyday household
  tasks,'' \emph{arXiv preprint arXiv:2108.03272}, 2021.

\bibitem{wake2021}
N.~Wake, R.~Arakawa, I.~Yanokura, T.~Kiyokawa, K.~Sasabuchi, J.~Takamatsu, and
  K.~Ikeuchi, ``A learning-from-observation framework: One-shot robot teaching
  for grasp-manipulation-release household operations,'' in \emph{2021
  IEEE/SICE International Symposium on System Integration (SII)}, 2021, pp.
  461--466.

\bibitem{kazhoyan2021}
G.~Kazhoyan, S.~Stelter, F.~K. Kenfack, S.~Koralewski, and M.~Beetz, ``The
  robot household marathon experiment,'' in \emph{2021 IEEE International
  Conference on Robotics and Automation (ICRA)}, 2021, pp. 9382--9388.

\bibitem{yolov8_ultralytics}
\BIBentryALTinterwordspacing
G.~Jocher, A.~Chaurasia, and J.~Qiu, ``Ultralytics yolov8,'' 2023. [Online].
  Available: \url{https://github.com/ultralytics/ultralytics}
\BIBentrySTDinterwordspacing

\bibitem{kirillov2023segment}
A.~Kirillov, E.~Mintun, N.~Ravi, H.~Mao, C.~Rolland, L.~Gustafson, T.~Xiao,
  S.~Whitehead, A.~C. Berg, W.-Y. Lo \emph{et~al.}, ``Segment anything,''
  \emph{arXiv preprint arXiv:2304.02643}, 2023.

\bibitem{takata2022graph}
K.~Takata, T.~Kiyokawa, N.~Yamanobe, I.~G. Ramirez-Alpizar, W.~Wan, and
  K.~Harada, ``Graph-based framework on bimanual manipulation planning from
  cooking recipe,'' \emph{Robotics}, vol.~11, no.~6, p. 123, 2022.

\bibitem{sakib2022cooking}
M.~S. Sakib, D.~Paulius, and Y.~Sun, ``Approximate task tree retrieval in a
  knowledge network for robotic cooking,'' \emph{IEEE Robotics and Automation
  Letters}, vol.~7, no.~4, pp. 11\,492--11\,499, 2022.

\bibitem{delpreto2022actionsense}
J.~DelPreto, C.~Liu, Y.~Luo, M.~Foshey, Y.~Li, A.~Torralba, W.~Matusik, and
  D.~Rus, ``Actionsense: A multimodal dataset and recording framework for human
  activities using wearable sensors in a kitchen environment,'' \emph{Advances
  in Neural Information Processing Systems}, vol.~35, pp. 13\,800--13\,813,
  2022.

\bibitem{zhang2019}
K.~Zhang, M.~Sharma, M.~Veloso, and O.~Kroemer, ``Leveraging multimodal haptic
  sensory data for robust cutting,'' in \emph{2019 IEEE-RAS 19th International
  Conference on Humanoid Robots (Humanoids)}, 2019, pp. 409--416.

\bibitem{redmon2016you}
J.~Redmon, S.~Divvala, R.~Girshick, and A.~Farhadi, ``You only look once:
  Unified, real-time object detection,'' in \emph{Proceedings of the IEEE
  conference on computer vision and pattern recognition}, 2016, pp. 779--788.

\bibitem{redmon2017yolo9000}
J.~Redmon and A.~Farhadi, ``Yolo9000: better, faster, stronger,'' in
  \emph{Proceedings of the IEEE conference on computer vision and pattern
  recognition}, 2017, pp. 7263--7271.

\bibitem{redmon2018yolov3}
------, ``Yolov3: An incremental improvement,'' \emph{arXiv preprint
  arXiv:1804.02767}, 2018.

\bibitem{bochkovskiy2020yolov4}
A.~Bochkovskiy, C.-Y. Wang, and H.-Y.~M. Liao, ``Yolov4: Optimal speed and
  accuracy of object detection,'' \emph{arXiv preprint arXiv:2004.10934}, 2020.

\bibitem{norizan2023object}
N.~A.~A. Norizan, M.~R.~M. Tomari, and W.~N.~W. Zakaria, ``Object detection
  using yolo for quadruped robot manipulation,'' \emph{Evolution in Electrical
  and Electronic Engineering}, vol.~4, no.~1, pp. 329--336, 2023.

\bibitem{mou2022pose}
F.~Mou, H.~Ren, B.~Wang, and D.~Wu, ``Pose estimation and robotic insertion
  tasks based on yolo and layout features,'' \emph{Engineering Applications of
  Artificial Intelligence}, vol. 114, p. 105164, 2022.

\bibitem{radford2021learning}
A.~Radford, J.~W. Kim, C.~Hallacy, A.~Ramesh, G.~Goh, S.~Agarwal, G.~Sastry,
  A.~Askell, P.~Mishkin, J.~Clark \emph{et~al.}, ``Learning transferable visual
  models from natural language supervision,'' in \emph{International conference
  on machine learning}.\hskip 1em plus 0.5em minus 0.4em\relax PMLR, 2021, pp.
  8748--8763.

\bibitem{shridhar22a}
\BIBentryALTinterwordspacing
M.~Shridhar, L.~Manuelli, and D.~Fox, ``Cliport: What and where pathways for
  robotic manipulation,'' in \emph{Proceedings of the 5th Conference on Robot
  Learning}, ser. Proceedings of Machine Learning Research, A.~Faust, D.~Hsu,
  and G.~Neumann, Eds., vol. 164.\hskip 1em plus 0.5em minus 0.4em\relax PMLR,
  08--11 Nov 2022, pp. 894--906. [Online]. Available:
  \url{https://proceedings.mlr.press/v164/shridhar22a.html}
\BIBentrySTDinterwordspacing

\bibitem{nair2022r3m}
S.~Nair, A.~Rajeswaran, V.~Kumar, C.~Finn, and A.~Gupta, ``R3m: A universal
  visual representation for robot manipulation,'' \emph{arXiv preprint
  arXiv:2203.12601}, 2022.

\bibitem{grauman2022ego4d}
K.~Grauman, A.~Westbury, E.~Byrne, Z.~Chavis, A.~Furnari, R.~Girdhar,
  J.~Hamburger, H.~Jiang, M.~Liu, X.~Liu \emph{et~al.}, ``Ego4d: Around the
  world in 3,000 hours of egocentric video,'' in \emph{Proceedings of the
  IEEE/CVF Conference on Computer Vision and Pattern Recognition}, 2022, pp.
  18\,995--19\,012.

\bibitem{wang2023manipulate}
J.~Wang, S.~Dasari, M.~K. Srirama, S.~Tulsiani, and A.~Gupta, ``Manipulate by
  seeing: Creating manipulation controllers,'' \emph{arXiv preprint
  arXiv:2303.08135}, 2023.

\bibitem{chane2023learning}
E.~Chane-Sane, C.~Schmid, and I.~Laptev, ``Learning video-conditioned policies
  for unseen manipulation tasks,'' \emph{arXiv preprint arXiv:2305.06289},
  2023.

\bibitem{ikegami2020}
N.~Ikegami, S.~Arnold, K.~Nagahama, and K.~Yamazaki, ``Active learning of the
  cutting of cooking ingredients using simulation with object splitting,'' in
  \emph{2020 IEEE/SICE International Symposium on System Integration (SII)},
  2020, pp. 1--8.

\bibitem{sawhney2021playing}
A.~Sawhney, S.~Lee, K.~Zhang, M.~Veloso, and O.~Kroemer, ``Playing with food:
  Learning food item representations through interactive exploration,'' in
  \emph{Experimental Robotics: The 17th International Symposium}.\hskip 1em
  plus 0.5em minus 0.4em\relax Springer, 2021, pp. 309--322.

\bibitem{blodow2011}
N.~Blodow, L.~C. Goron, Z.-C. Marton, D.~Pangercic, T.~Rühr, M.~Tenorth, and
  M.~Beetz, ``Autonomous semantic mapping for robots performing everyday
  manipulation tasks in kitchen environments,'' in \emph{2011 IEEE/RSJ
  International Conference on Intelligent Robots and Systems}, 2011, pp.
  4263--4270.

\bibitem{zhu2017}
Q.~Zhu, V.~Perera, M.~Wächter, T.~Asfour, and M.~Veloso, ``Autonomous
  narration of humanoid robot kitchen task experience,'' in \emph{2017 IEEE-RAS
  17th International Conference on Humanoid Robotics (Humanoids)}, 2017, pp.
  390--397.

\bibitem{wang2020too}
R.~E. Wang, S.~A. Wu, J.~A. Evans, J.~B. Tenenbaum, D.~C. Parkes, and
  M.~Kleiman-Weiner, ``Too many cooks: Coordinating multi-agent collaboration
  through inverse planning,'' in \emph{Proceedings of the 19th International
  Conference on Autonomous Agents and MultiAgent Systems}, 2020, pp.
  2032--2034.

\bibitem{franka2021}
\emph{Franka Emika Robot's Instruction Handbook}.\hskip 1em plus 0.5em minus
  0.4em\relax Franka Emika GmbH, 2021.

\bibitem{zhang2020modular}
K.~Zhang, M.~Sharma, J.~Liang, and O.~Kroemer, ``A modular robotic arm control
  stack for research: Franka-interface and frankapy,'' \emph{arXiv preprint
  arXiv:2011.02398}, 2020.

\end{thebibliography}

\end{document}